\documentclass[letterpaper]{article} 
\usepackage{aaai24}  
\usepackage{times}  
\usepackage{helvet}  
\usepackage{courier}  
\usepackage[hyphens]{url}  
\usepackage{graphicx} 
\usepackage{natbib}  
\usepackage{caption} 
\usepackage{longtable} 
\usepackage{algorithm}
\usepackage{algorithmic}

\usepackage{newfloat}
\usepackage{listings}
\DeclareCaptionStyle{ruled}{labelfont=normalfont,labelsep=colon,strut=off} 
\floatstyle{ruled}
\newfloat{listing}{tb}{lst}{}
\floatname{listing}{Listing}

\usepackage{amsmath}
\usepackage{booktabs}

\usepackage{array}        
\usepackage{placeins}     

\title{Evaluating the Efficacy of LLMs to Emulate Realistic Human Personalities}
 \author {
     Lawrence J. Klinkert,
     Steph Buongiorno,
     Corey Clark
 }
 \affiliations {
     Southern Methodist University\\
     Jklinkert@mail.smu.edu, sbuongiorno@smu.edu, coreyc@mail.smu.edu
}

\begin{document}

\maketitle

\begin{abstract}
To enhance immersion and engagement in video games, the design of Affective Non-Player Characters (ANPCs) is a key focus for researchers and practitioners. Affective Computing frameworks improve Non-player characters (NPC) by providing personalities, emotions, and social relations. Large Language Models (LLMs) bring the promise to dynamically enhance character design when coupled with these frameworks, but further research is needed to validate the models truly represent human qualities. In this research, a comprehensive analysis investigates the capabilities of LLMs to generate content that aligns with human personality, using the Big Five and human responses from the International Personality Item Pool (IPIP) questionnaire. Our goal is to benchmark the performance of various LLMs, including frontier models and local models, against an extensive dataset comprising over 50,000 human surveys of self-reported personality tests to determine whether LLMs can replicate human-like decision-making with personality-driven prompts. A range of personality profiles were used to cluster the test results from the human survey dataset. Our methodology involved prompting LLMs with self-evaluated test items for each personality profile, comparing their outputs to human baseline responses, and evaluating the accuracy and consistency. Our findings show that some local models had 0\% alignment of any personality profiles when compared to the human dataset, while the frontier models, in some cases, had 100\% alignment. The results indicate that NPCs can successfully emulate human-like personality traits using LLMs, as demonstrated by benchmarking the LLM's output against human data.  This foundational work serves as a methodology for game developers and researchers to test and evaluate LLMs, ensuring they accurately represent the desired human personalities and can be expanded for further validation.
\end{abstract}

\section{Introduction}
In the profession of video game design, there is broad interest in incorporating human-like personality into the design of Non-Playable Characters (NPCs) as a way to drive NPCs' behavior and emotional expression \cite{isbister_better_2006, durupinar_psychological_2016, rabin_personality_2015, jenner_fast_2017, boeda_driving_2021}. NPCs driven by personality models, such as the Big Five, can interact with the environment while displaying character depth and emotional complexity, which can improve player engagement \cite{ isbister_better_2006, klinkert_artificial_2021, shirvani_personality_2023}. However, designing NPCs with human-like personality traits poses significant challenges, such as recognizing and expressing these traits, integrating them with game mechanics and narratives, and ensuring consistent behavior that allows for replayability \cite{ durupinar_psychological_2016, garavaglia_moody5_2022, shirvani_personality_2023}.  Current NPC design often fails to meet these objectives, resulting in a gaming experience that is less immersive and engaging \cite{ isbister_better_2006, nieweglowska_ticket_2024}. Affective Computing provides techniques to improve NPC design.

Affective Computing (AC) uses psychological models to create human-like behaviors in NPCs, enhancing their personality, emotions, and social interactions in games \cite{durupinar_psychological_2016, klinkert_artificial_2021, shirvani_personality_2023}. This approach anthropomorphizes digital characters by attributing human traits to them. For instance, an NPC's actions might be described as "happy when completing a task" or "anxious about the party" \cite{klinkert_artificial_2021, shirvani_personality_2023}. However, these systems need the means to display these traits, either through facial expressions, voice editing, or descriptive text \cite{bidarra_growing_2010, schiffer_facial_2022}. Large Language Models (LLMs) can supplement AC systems to enhance NPC believability by enabling human-like personality traits and dynamic context-aware interactions.

This research demonstrates that prompting an LLM with psychometric values derived from an AC system can generate content that better aligns with the expected behavior of an NPC’s personality. This study uses the International Personality Item Pool (IPIP) 50 questionnaire to measure human personality traits. The questionnaire is a recognized tool in psychological research for assessing the Big Five personality traits, making it an essential component for ensuring the validity of our benchmarks \cite{goldberg_broad-bandwidth_1999}.

Our methodology involves prompting various LLMs, such as OpenAI's GPT models, Google's flan models, and Mistral's local models, with personality questions from the IPIP-50 questionnaire and comparing their synthetic responses to human responses. These models can be categorized into two distinct types: "frontier" models, which are large-scale models with over 100 billion parameters, and "local" models, which are smaller, task-optimized models with under 100 billion parameters. Frontier models like GPT-4 are designed for general-purpose tasks and leverage vast amounts of data and computational resources, making them highly versatile but resource-intensive. In contrast, local models like Dolphin-2.7-Mistral are optimized for specific tasks, offering a more cost-effective solution with a narrower focus but lower overall capacity. By analyzing how well the LLMs' outputs match the expected personality traits, we demonstrate that different LLMs can emulate human personalities to varying degrees. 

To validate our research, we cross-examine the LLMs' responses with human data, ensuring that our analyses are grounded in actual human behavior. Our results indicate significant variability among models: some local models demonstrated 0\% alignment with human personality profiles, whereas frontier models, like OpenAI’s GPT-4, exhibited a remarkable 100\% consistency in matching human responses. This success suggests that additional psychometric values can further refine LLM-generated content to align with an NPC’s psyche.

The following sections will explore our methodology in greater detail. Section two will elucidate the background information, encompassing topics such as video game companies researching AC for their games, the latest scholarly research in AC, LLM uses in video games, and the rationale for integrating AC with an LLM. Section three will delve into personality psychometrics, specifically the Big Five personality model. Section four will elaborate on the dataset of human responses to the IPIP-50 and how its information can be used to benchmark against an LLM. Section five will cover the generation process of synthetic data from the LLM and compare its results against the human baseline dataset. Section six will outline a use case of an LLM using personality for gameplay. Section seven will list the limitations of this work. Section eight, the conclusion, will summarize the results and point to future research.

\section{Background}
Industry practitioners and researchers increasingly endeavor to make more compelling NPCs through various means, including integrating AC systems into their games. Industry game studios such as Square Enix, Worldwalker Games, and Eidos Sherbrooke are exploring the possibilities of AC and its assimilation into their gaming products \cite{boeda_ai_2021, austin_independent_2022, trachel_machine_2022}. These corporations employ AC to empower NPCs to emulate emotions within their games. In executing this approach, these video game corporations experimented with emotive simulations for NPCs so that a player can recognize and empathize with the character. As for researchers, AC systems such as Popescu et al.'s GAMYGDALA, Shirvani \& Ware's emotional narrative planning system, and Klinkert and Clark's Artificial Psychosocial Framework (APF), are proficient in discerning, interpreting, processing, and simulating human emotional states between player and NPCs \cite{popescu_gamygdala_2014, shirvani_formalization_2020, klinkert_artificial_2021}. Integrating LLMs enhances this effort by improving the sophistication of NPC dialogue generation and decision-making in interactions with players and other NPCs.

The advent of LLMs is transformative in today's technological landscape, with companies like Microsoft, Google, Meta, and Nvidia integrating them into their products \cite{zhao_survey_2023}. However, LLMs face challenges such as data hallucinations, memory limitations, and restricted accessibility \cite{azamfirei_large_2023, wang_augmenting_2023}. These challenges are particularly pronounced in the video game industry, where LLMs are increasingly used for dynamic content generation. The inherent unpredictability of LLMs makes it difficult to ensure that in-game characters consistently behave and express themselves in ways that align with their design. This inconsistency can disrupt player immersion, requiring extensive retesting to maintain the unique and engaging experiences that modern games strive to deliver \cite{takeyama_unexpected_2023}. These complexities and challenges have made the industry hesitant to embrace LLMs fully \cite{millington_artificial_2019}.

Nevertheless, one exemplary study by Park et al. demonstrates the integration of an LLM into a simulated village, showing that NPCs can understand their surroundings, plan their activities, and share information with other NPCs  \cite{park_generative_2023}. This integration enhances NPCs' believability. However, to ensure that the content aligns with the NPC's personality and emotional state, an AC system can provide psychometric values that can be integrated with LLMs, pushing these characters further toward human-like behavior.

AC systems provide psychometric values, including personality traits, emotional states, and relationship dynamics between game entities \cite{popescu_gamygdala_2014, shirvani_formalization_2020, klinkert_artificial_2021}. This paper explores whether an LLM can effectively interpret and interact with these values, initially focusing on personality. Personality traits, based on models like the Big Five, significantly impact a character's decisions, actions, and reactions within a game's narrative. Thus, an LLM's ability to exemplify these traits is crucial for creating believable characters with consistent behavior.

AC systems capitalize on personality models such as the Big Five, or the Five Factor or OCEAN model, to represent a character's fundamental psychological composition within a game \cite{klinkert_artificial_2021}. This model comprises five diverse factors: \textbf{O}penness to novel experiences, \textbf{C}onscientiousness in tasks and interpersonal relationships, \textbf{E}xtraversion in social contexts, \textbf{A}greeableness towards diverse viewpoints and mutual understandings, and \textbf{N}euroticism in interpreting circumstances \cite{ john_big_1999}. These traits define a character's personality and influence how they respond to different scenarios for personality assessments and in-game decisions.

We propose that integrating a personality model with an LLM can enhance prompt optimization, apply general knowledge more broadly, and enable dynamic character transformations in video games. By representing the Big Five personality traits as a 5-tuple structure—each trait corresponding to one of the dimensions (Openness, Conscientiousness, Extraversion, Agreeableness, Neuroticism)—we can reduce the number of tokens needed to convey the expected personality for an NPC. This simplified structure makes each trait distinct and easy to represent, facilitating more efficient communication with the model and ensuring that the character's personality is consistently reflected in their behavior and dialogue.

Given the extensive array of internet-based data used to train an LLM, it is plausible that research related to the Big Five is part of that dataset, including examples of individuals exemplifying each personality factor \cite{karra_estimating_2023, jiang_evaluating_2024}. Research has shown that LLMs can express certain personality traits and perform self-evaluated surveys when prompted to adopt a specific personality \cite{serapio-garcia_personality_2023, hagendorff_machine_2023, song_have_2023}. These studies use human psychological methodologies to validate their work but have not yet used human data as a baseline. Our research extends their work by using human data as a baseline, ensuring that our approach is grounded in actual human behavior.

Recent studies suggest that responding to personality questionnaires involves a decision-making process analogous to the choices made in real-world scenarios. For instance, Karra et al. found that the traits exhibited by LLMs during these tests reflect a form of decision-making where the model evaluates which response best aligns with its learned personality traits \cite{karra_estimating_2023}. Similarly, Lee et al. argues that the consistency in LLM responses across different prompts indicates an internal evaluation process akin to decision-making \cite{lee_llms_2024}. Thus, the idea that answering personality questionnaires is not merely a selection task but a microcosm of broader decision-making processes, which can be extended to the context of video game interactions.

\begin{figure}
    \centering
    \includegraphics[width=\linewidth]{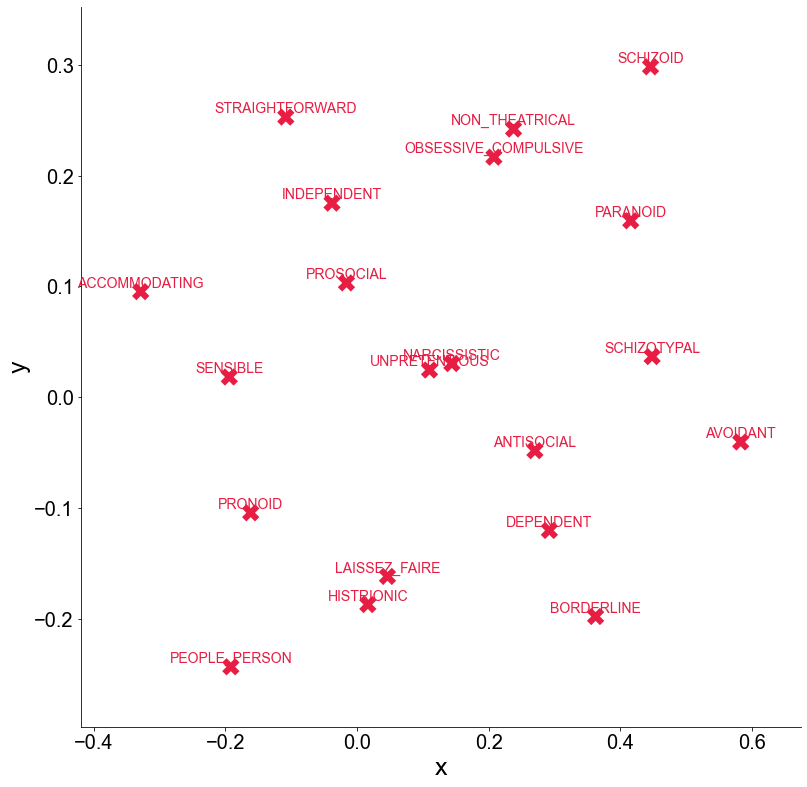}
    \caption{Van Mensvoort's 20 personality Profiles, plotted via PCA. \cite{mensvoort_system_nodate}.}
    \label{fig:Van}
\end{figure}

\section{Personality Representation}
The Big Five represents different characteristics into five categories: \textbf{O}penness, \textbf{C}onscientiousness, \textbf{E}xtraversion, \textbf{A}greeableness, and \textbf{N}euroticism, which are represented on a continuous scale ranging from zero to one. A score of one signifies the full expression of a particular trait, while zero implies the presence of the opposing attribute \cite{goldberg_development_1992}.

A score of 1.0 on the Openness dimension indicates high creativity, a readiness to embrace novelty, a drive to tackle new challenges, and engagement in abstract thought, while a score of 0.0 denotes resistance to change, a lack of interest in new experiences, unwillingness to embrace new ideas, and a shortage of creative thought. Thus, the Openness factor can be viewed as a spectrum from cautious/consistent to inventive/curious. Importantly, values assigned to any trait are not restricted to extremes but can occupy any position within this range, offering a nuanced representation of personality traits. This model spans five dimensions with infinitesimal graduations, enabling a virtually limitless variety of personality combinations across these factors.

Drawing on the work of Van Mensvoort, the Big Five can be discretized into 20 distinct personality profiles \cite{mensvoort_system_nodate}. This subdivision encompasses a range of behavioral characteristics, from behavioral disorders such as Paranoid and Schizoid to their counterparts, Pronoid and People-person. This fine-grained classification, shown in Figure \ref{fig:Van}, allows for a more precise and detailed depiction of personality, accommodating its complex, multifaceted nature.

Employing these profiles as categorical labels, we leverage the human data to establish our Baseline in this research. This Baseline serves as a comparative standard against which we evaluate the performance of our methodology and the LLM's capacity to represent and express various character personalities. Thus, the defined profiles and our chosen baseline data inform our research approach and assist in objectively evaluating our outcomes. 

\begin{table*}
  \centering
  \begin{tabular}{l||rrrrr}
    \toprule
    Baseline & Openness & Conscientiousness & Extraversion & Agreeableness & Neuroticism \\
    \midrule
      count & 50500 & 50500 & 50500 & 50500 & 50500 \\
        mean & 0.699831 & 0.584225 & 0.449248 & 0.566697 & 0.512838 \\
        std & 0.177709 & 0.205618 & 0.242912 & 0.219459 & 0.230343\\
        min & 0 & 0 & 0 & 0 & 0 \\
        25\% & 0.575 & 0.425 & 0.25 & 0.4 & 0.35 \\
        50\% & 0.725 & 0.575 & 0.45 & 0.575 & 0.525 \\
        75\% & 0.825 & 0.75 & 0.625 & 0.725 & 0.675\\
        max & 1 & 1 & 1 & 1 & 1 \\
    \bottomrule
  \end{tabular}
  \caption{The statistical description of the evaluated test results from the IPIP Big Five Factor Markers dataset.}
  \label{Table:describe}
\end{table*}

\section{Data Collection}

We investigate the results from a comprehensive Big Five personality dataset using the personality profiles. The dataset, derived from the Open-Source Psychometrics Project, comprises personality test results from over a million participants (n = 1,015,342) who responded to the IPIP-50 \cite{noauthor_open_nodate, goldberg_development_1992}. Each test item was a statement to which participants responded on a Likert scale, with one indicating a strong disagreement with the presented behavior and five denoting strong agreement.

We undertook a thorough data-cleaning process to optimize the dataset for our research. We first ensured that each test response included answers to all 50 statements. We further filtered the data to include only those responses that took more than 300 milliseconds to answer a question, as the average human response time ranges between 150 and 300 milliseconds, thus mitigating automated submissions to the analysis. Finally, we only retained the first submission from each unique IP address. Following these preparation steps, a total of 596,956 test results remained for our analysis. This cleaned dataset now forms the foundation for our subsequent investigations. The full and clean versions of the data set are accessible in our public repository \footnote{\url{https://gitlab.com/humin-game-lab/llm-personality}}.

The next phase involved evaluating the personality test results using the scoring key from the IPIP website \cite{goldberg_ipip_nodate}. Each test statement corresponds to a personality factor, and participants' Likert values are totaled for each factor, with inversions applied for negatively worded statements. For example, a response of 5 to "Doesn't talk a lot" (reflecting low Extraversion) is inverted to 1. Scores for each personality factor range from 10 to 50 and are normalized to a [0.0, 1.0] scale using linear interpolation. The final output is a 5-tuple vector (O, C, E, A, N), accurately representing an individual's personality configuration based on the Big Five model.

We executed the label assignment of a personality profile with the evaluated test results using a nearest-neighbor approach. For every test result represented as a 5-tuple vector, we calculated the Euclidean Distance to each of the 20 personality profiles. The personality profile with the smallest distance to the test results was identified as the corresponding label. 

The last step in our preparation involved disproportionate sampling for each personality profile. Our research does not seek to replicate the distribution of personalities in the human population. Instead, we employ disproportionate sampling to ensure an equal number of samples for each personality profile. This approach allows us to fairly compare the profiles without one dominating another, ensuring that we adequately represent each subpopulation to construct the general shape of the population while mitigating the influence of outliers that might otherwise blur the boundaries between different personality profile clusters. Consequently, our baseline dataset incorporated a total of 50,500 test results. The statistical description can be found in Table \ref{Table:describe}. The subset version of the cleaned dataset can also be found in the provided repository.

To understand the complex interplay of five-dimensional personality data, we utilize Principal Component Analysis (PCA), a recognized method for reducing dimensionality \cite{wold_principal_1987, abdi_principal_2010, greenacre_principal_2022}. PCA transforms high-dimensional data by identifying the principal components, which are the directions of maximum variance in the data. It then projects the data points onto these principal components, effectively reducing the dimensionality while preserving as much variance as possible. Figure \ref{fig:Van} shows the two-dimensional representation of the personality profiles.

\begin{figure}
  \centering
  \includegraphics[width=1.0\linewidth]{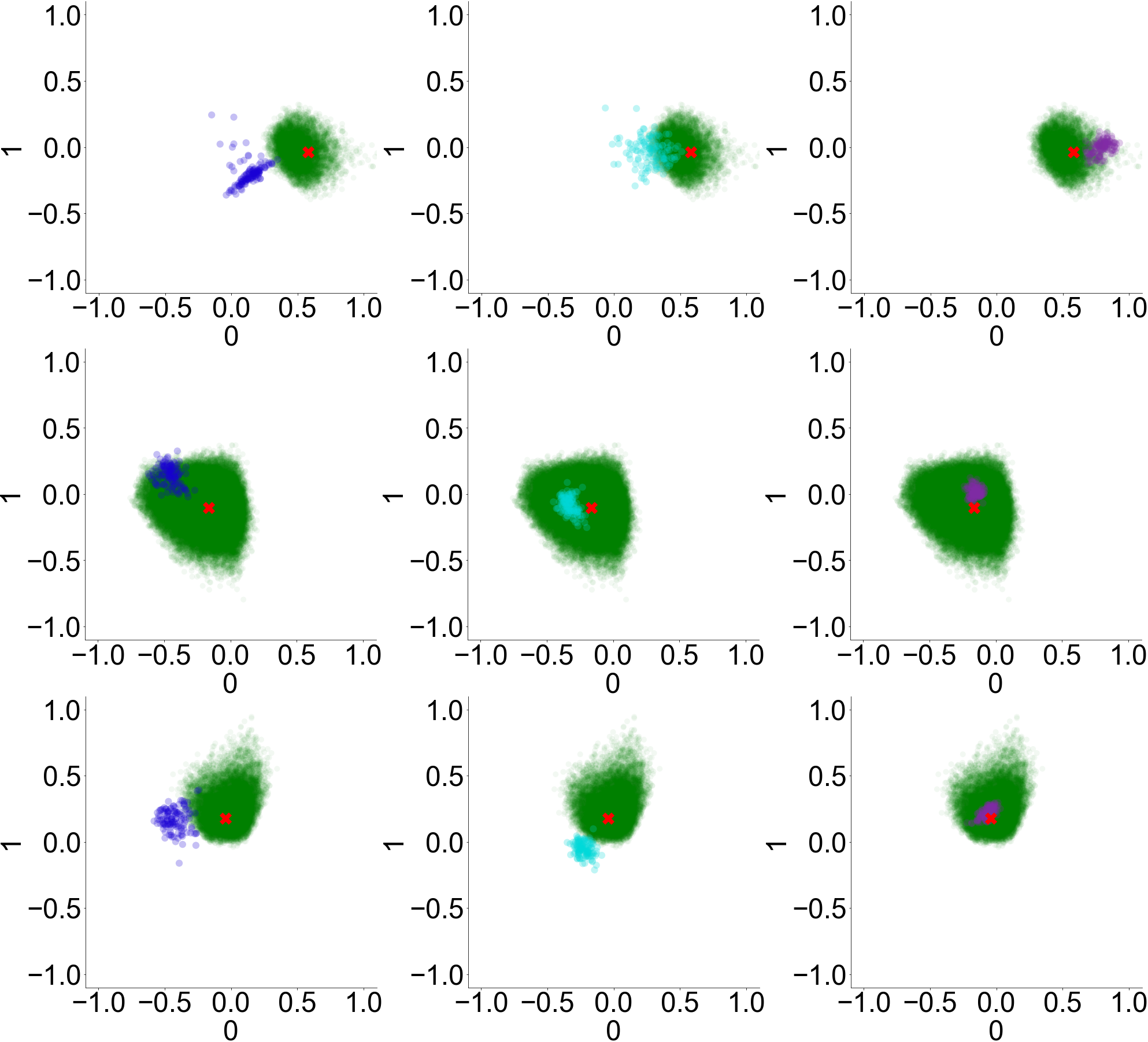}
  \caption{Nine different plots of evaluated test results using PCA. The green circles represent the human baseline results. Each row is a different personality profile: Avoidant, Pronoid, Independent, each represented by a red X. Each column is a different LLM test result, starting with GPT davinci, GPT-3.5, then GPT-4, represented by blue, cyan, and purple circles, respectively.}
  \label{fig:compare}
\end{figure}

Figure \ref{fig:compare} shows a few examples of human test results forming clusters around the personality profiles, represented by the central X within each cluster. From the top row of graphs to the bottom row, the profiles represent Avoidant, Pronoid, and Independent. From Left to right, we plot the synthetic results created by LLMs, specifically GPT Davinci, GPT-3.5, and GPT-4. The next section will discuss how the synthetic data was generated.

\section{Generate Synthetic Data}
Several frontier and local models were selected for this study, including Flan-XL, Flan-XXL, Text-DaVinci-003, GPT-3.5-Turbo-0613, GPT-4-0613, Dolphin-2.7-Mistral, and Mixtral-8x7B-Instruct. Using a diverse set of models was essential to ensure the results were not model-specific and to evaluate the consistency of personality traits across different architectures and training methodologies. Each model was presented with the self-assessment questions in a randomized order, with each question prompted separately. This approach follows the recommendations by Song et al. and Jiang et al., who highlight the importance of minimizing context-induced biases and ensuring reliable responses through randomization and isolated prompting \cite{song_have_2023, jiang_evaluating_2024}. By adhering to these best practices, we aimed to achieve a robust and unbiased assessment of LLM-generated personality traits.

To illustrate, a prompt designed for a "laissez-faire" personality profile reads as follows:

\begin{quote}
"For the following task, I want you to pretend to be a human who responds in a way that matches this Five Factor Personality: (O:0.7315, C:0.1455, E:0.6305, A:0.5295, N:0.4375). Considering the statement, 'Get upset easily.', please rate how accurately this describes you on a scale from 1 to 5 (where 1 = 'very inaccurate', 2 = 'moderately inaccurate', 3 = 'neither accurate nor inaccurate', 4 = 'moderately accurate', and 5 = 'very accurate'). Respond with 'I would rate this as X (Likert rating description).'".
\end{quote}

Serapio-Garcia et al. and Song et al. advocate for multiple iterations of the personality tests to achieve reliable results \cite{serapio-garcia_personality_2023, song_have_2023}. Thus, each model was assigned one of the 20 target personality profiles during the test, and the process was repeated 64 times for each profile, ensuring statistical significance. This comprehensive approach allowed for a thorough evaluation of each models' ability to simulate diverse personality traits.

\begin{table}
  \centering
  \begin{tabular}{l|rr}
    \toprule
    Model & Accuracy&95\% CI\\
    \midrule
     flan-xl & 5.00\% &  $\pm$ 4.628\%\\
     flan-xxl & 5.00\% & $\pm$ 1.002\%\\
     text-davinci-003 & 15.10\% & $\pm$ 1.164\% \\
     GPT-3.5-turbo-0613 & 31.90\% & $\pm$ 1.516\% \\
     GPT-4-0613 & 74.00\% & $\pm$ 1.426\% \\
     dolphin-2.7-mixtral & 7.11\% & $\pm$ 1.186\% \\
     mixtral-8x7B-instruct & 6.33\% & $\pm$ 1.120\% \\
    \bottomrule
  \end{tabular}
  \caption{LLM Accuracy of Personality Profile prompting and test evaluation.}
  \label{tab:accuracy}
\end{table}

\begin{table}[ht!]
    \centering
    \small
    \begin{tabular}{lp{1.5cm}rrrrr}
        \toprule
         Model & Profile & O & C & E & A & N \\
        \midrule
        GPT 4                     & Non theatrical       & 0.06 & 0.08 & 0.02 & 0.11 & 0.04 \\
        GPT 4                     & Pronoid              & 0.09 & 0.08 & 0.07 & 0.03 & 0.06 \\
        GPT 4                     & Avoidant             & 0.05 & 0.10 & 0.04 & 0.10 & 0.04 \\
        GPT 4                     & Borderline           & 0.09 & 0.03 & 0.09 & 0.05 & 0.09 \\
        GPT 3.5                   & Unpretentious        & 0.10 & 0.05 & 0.07 & 0.08 & 0.05 \\
        GPT 4                     & Schizoid             & 0.02 & 0.10 & 0.04 & 0.10 & 0.08 \\
        GPT 3.5                   & Prosocial            & 0.13 & 0.06 & 0.05 & 0.04 & 0.06 \\
        GPT 3.5                   & Non theatrical       & 0.14 & 0.04 & 0.05 & 0.04 & 0.06 \\
        ...                       & ...                  & ...    & ...    & ...    & ...    & ...    \\
        dolphine                  & Avoidant             & 0.19 & 0.08 & 0.34 & 0.17 & 0.50 \\
        davinci                   & Laissez faire        & 0.13 & 0.59 & 0.08 & 0.23 & 0.17 \\
        davinci                   & Narcissistic         & 0.26 & 0.29 & 0.09 & 0.51 & 0.19 \\
        davinci                   & Avoidant             & 0.42 & 0.33 & 0.29 & 0.29 & 0.09 \\
        davinci                   & Obsessive compulsive & 0.58 & 0.02 & 0.16 & 0.29 & 0.20 \\
        davinci                   & Antisocial           & 0.19 & 0.33 & 0.17 & 0.48 & 0.31 \\
        davinci                   & Schizoid             & 0.46 & 0.32 & 0.31 & 0.34 & 0.24 \\
        davinci                   & Paranoid             & 0.44 & 0.24 & 0.12 & 0.56 & 0.19 \\
        \bottomrule
    \end{tabular}
    \caption{The MSPE of each personality profile based on Euclidean Distance (ED), sorted top to bottom from closest to farthest}
    \label{tab:MSPE}
\end{table}

\subsection{Synthetic Data and Personality Labels}
Our analysis evaluates the accuracy of the LLM's results by comparing them to the assigned personality profiles. This assessment involves calculating the shortest Euclidean distance between the test results and each personality profile, using the same approach as the baseline dataset. Accuracy is measured by the frequency with which the assigned personality profile is also the closest profile to the generated result. The resulting accuracies highlight significant differences in each model's performance, as shown in Table \ref{tab:accuracy}. The flan models performed the poorest, with an accuracy of 5.0\% ($\pm$ 4.6\% and $\pm$ 1.0\% at 95\% confidence interval), followed by the mixtral models at 6.33\% ($\pm$ 1.1\%  at 95\% confidence interval) and 7.11\% ($\pm$ 1.2\%  at 95\% confidence interval). In contrast, the next-generation model, GPT-4-0613, exhibited a substantial leap in performance, achieving an accuracy of 74.0\% ($\pm$ 1.4\% at 95\% confidence interval).
  	 
To determine how far off the model's prediction was, we utilized the Mean Squared Prediction Error (MSPE). This measurement assumes that if the personality profile's labeling is accurate, the error in the evaluated test results should be marginal; thus, they should closely align with the respective profile. Table \ref{tab:MSPE} shows a subset of the full table, revealing the top profiles that were accurately represented and the least accurate representations. The full table can be reviewed in our repository. The Extroverted trait was well-represented in all the models, followed by Conscientiousness, Openness, Agreeableness, and Neuroticism. GPT 4 had the best test results and was consistently close to representing the original personality label. The profiles well represented by all models were Prosocial, Unpretentious, and Pronoid, while laissez-faire, Paranoid, and Avoidant were the least representative.

\begin{table*}
    \centering
    \begin{tabular}{l||rrrrrrr}
    \toprule
    & flan xl & flan xxl & davinci & GPT 3.5 & GPT 4 & dolphine mixtral & mistral instruct \\
    \midrule
Paranoid             & 0.0000   & 0.0000    & 0.0000  & 0.0547   & 0.6563 & 0.0000                     & 0.0000                     \\
Schizoid             & 0.0000   & 0.0000    & 0.0000  & 0.1563   & 0.8203 & 0.0000                     & 0.0000                     \\
Schizotypal          & 0.0000   & 0.0000    & 0.0078  & 0.3672   & 0.5547 & 0.0000                     & 0.0000                     \\
Antisocial           & 0.0000   & 0.0000    & 0.0000  & 0.0000   & 0.0938 & 0.0000                     & 0.0000                     \\
Borderline           & 0.0000   & 0.0000    & 0.0000  & 0.1875   & 0.9922 & 0.0000                     & 0.0000                     \\
Histrionic           & 0.0000   & 0.0000    & 0.0000  & 0.0547   & 0.5547 & 0.0000                     & 0.0000                     \\
Narcissistic         & 0.0000   & 0.0000    & 0.0000  & 0.0000   & 0.1953 & 0.0000                     & 0.0000                     \\
Avoidant             & 0.0000   & 0.0000    & 0.0078  & 0.5000   & 0.9844 & 0.0000                     & 0.0000                     \\
Dependent            & 0.0000   & 0.0000    & 0.6328  & 0.3984   & 0.9844 & 0.0000                     & 0.0000                     \\
Obsessive-compulsive & 0.0000   & 0.0000    & 0.0000  & 0.0000   & 0.9531 & 0.0000                     & 0.0000                     \\
Pronoid              & 0.0000   & 0.0000    & 0.1797  & 0.6484   & 1.0000 & 0.3281                     & 0.1875                     \\
People person        & 0.0000   & 0.0000    & 0.4688  & 0.7734   & 1.0000 & 0.0000                     & 0.0000                     \\
Sensible             & 0.0000   & 0.0000    & 0.1172  & 0.3984   & 0.9219 & 0.0156                     & 0.0156                     \\
Prosocial            & 0.0000   & 1.0000    & 0.2891  & 0.8359   & 0.9766 & 0.6875                     & 0.6719                     \\
Straightforward      & 0.0000   & 0.0000    & 0.6641  & 0.3125   & 0.4297 & 0.3438                     & 0.2813                     \\
Non theatrical       & 0.0000   & 0.0000    & 0.0000  & 0.8125   & 0.9063 & 0.0156                     & 0.0000                     \\
Unpretentious        & 0.0000   & 0.0000    & 0.0000  & 0.5547   & 0.9609 & 0.0156                     & 0.0313                     \\
Accommodating        & 0.0000   & 0.0000    & 0.5938  & 0.1563   & 0.7656 & 0.0000                     & 0.0000                     \\
Independent          & 0.0000   & 0.0000    & 0.0000  & 0.0547   & 0.6406 & 0.0000                     & 0.0000                     \\
Laissez faire        & 0.0000   & 0.0000    & 0.0000  & 0.0000   & 0.1875 & 0.0000                     & 0.0000 \\                  
    \bottomrule
    \end{tabular}
  \caption{The percentage of synthetic points that were inside of the convex hull created by human data.}
  \label{tab:convex}
\end{table*}

\begin{table*}
    \centering
    \begin{tabular}{l||rrrrrrr}
    \toprule
    & flan xl & flan xxl & davinci & GPT 3.5 & GPT 4 & dolphine mixtral & mistral instruct \\
    \midrule
Paranoid             & 10.915   & 20.064    & 35.227  & 12.909   & 5.556  & 32.223                     & 32.753                     \\
Schizoid             & 38.017   & 53.216    & 57.851  & 7.500    & 4.822  & 40.012                     & 39.270                     \\
Schizotypal          & 13.302   & 13.074    & 18.466  & 6.121    & 7.435  & 14.150                     & 14.096                     \\
Antisocial           & 5.017    & 14.839    & 39.449  & 10.818   & 6.828  & 26.437                     & 25.344                     \\
Borderline           & 16.802   & 19.899    & 14.084  & 9.205    & 3.994  & 43.393                     & 42.367                     \\
Histrionic           & 9.495    & 9.479     & 32.099  & 7.994    & 4.709  & 15.257                     & 15.151                     \\
Narcissistic         & 9.634    & 18.849    & 49.660  & 16.222   & 7.847  & 38.506                     & 39.406                     \\
Avoidant             & 46.235   & 47.638    & 22.176  & 8.494    & 3.939  & 75.584                     & 75.237                     \\
Dependent            & 8.774    & 8.975     & 9.644   & 6.319    & 4.975  & 26.664                     & 26.682                     \\
Obsessive compulsive & 23.057   & 25.615    & 22.302  & 20.067   & 4.186  & 47.693                     & 47.119                     \\
Pronoid              & 17.477   & 7.810     & 6.845   & 3.923    & 2.123  & 8.175                      & 8.249                      \\
People person        & 21.114   & 9.045     & 6.855   & 3.287    & 1.170  & 22.467                     & 23.141                     \\
Sensible             & 10.541   & 12.459    & 8.109   & 4.345    & 1.593  & 14.067                     & 14.288                     \\
Prosocial            & 8.220    & 2.196     & 7.996   & 3.592    & 1.917  & 5.668                      & 5.439                      \\
Straightforward      & 21.734   & 30.735    & 4.303   & 6.093    & 3.383  & 6.408                      & 8.017                      \\
Non theatrical       & 11.071   & 27.504    & 49.103  & 3.113    & 4.695  & 23.926                     & 25.582                     \\
Unpretentious        & 12.672   & 6.968     & 28.478  & 4.115    & 1.975  & 11.021                     & 10.796                     \\
Accommodating        & 35.708   & 41.714    & 4.361   & 8.144    & 3.427  & 20.850                     & 22.083                     \\
Independent          & 7.435    & 10.127    & 18.842  & 5.980    & 4.428  & 9.906                      & 9.370                      \\
Laissez faire        & 26.863   & 28.118    & 43.308  & 20.468   & 4.962  & 26.883                     & 27.627 
 \\                  
    \bottomrule
    \end{tabular}
  \caption{The mean Mahalanobis distance, or average standard deviations away a synthetic point is from a human distribution of a personality profile.}
  \label{tab:Mahala}
\end{table*}

\begin{figure*}
  \includegraphics[width=.975\linewidth]{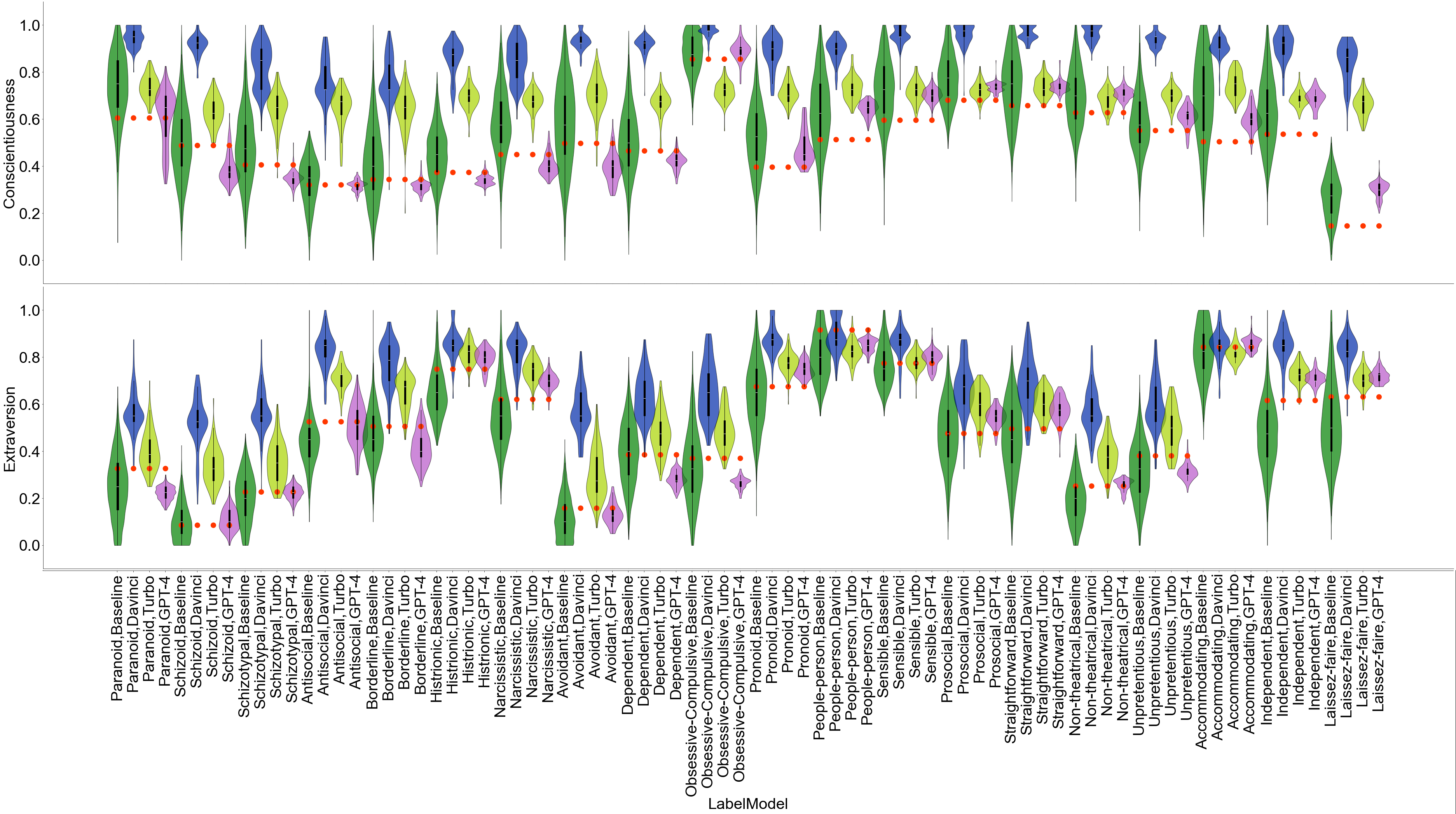}
  \centering
  \caption{Violin Plot of each personality profile (red point) against the baseline (green), davinci (blue), turbo (lime), and GPT-4 (purple) generated data evaluated test results along the Conscientiousness and Extraversion personality factors.}
  \label{fig:violin}
\end{figure*}

\subsection{Synthetic Data and Human Data}

To further support our data representing true personalities, we will compare the synthetic data generated by the LLMs with the human population. Our first measurement represented the percentage of synthetic data points within the cluster of human data. The second measurement considers the correlation between each dimension for a given personality profile and calculates an accurate distance between the synthetic point and the population.

To determine the percentage of synthetic data points inside a cluster of human data, we must define what it means to be inside a cluster. We took a direct approach and constructed a geometric shape from the human data, namely a convex hull. Our synthetic point is then defined as inside a cluster as long as it is within the convex hull, otherwise, the point is outside of the convex hull and is considered an outlier. 

As shown in Table \ref{tab:convex}, the Flan models could not represent the human data, except for its consistent guessing of Prosocial. Similarly, the local models of Dolphin and Mixtral were able to match Prosocial but struggled with several other profiles. Once again, the OpenAI models performed the best, with GPT 4 capable of having most of the results within the bounds of human data. Notice as the GPT model evolves from Davinci, GPT 3.5, and GPT 4, there was a high accuracy in the behavior wellness personality profiles, such as Dependent, Pronoid, and People Person, but the current generation model was capable of also representing the behavior disorder counterparts, Independent, Paranoid, and Schizoid.

For the next measurement, we needed to evaluate the similarity between the synthetic data and the human clusters. Therefore, we calculated the Mahalanobis distance in 5-dimensional space. The Mahalanobis distance measures how many standard deviations a point is from the mean of a distribution, considering the correlations between variables. This distance metric is particularly useful in multivariate spaces, where it accounts for the variances and covariances of the variables, providing an accurate measure of distance that reflects the true distribution of the data. 

The analysis of Mahalanobis distances in Table \ref{tab:Mahala} reveals that GPT-4 consistently outperforms other models, showing smaller distances and better alignment with the actual data distribution across various profiles. Conversely, Davinci and Mixtral models exhibit larger distances, indicating less accurate representations. Notably, profiles like Prosocial and Independent are well-represented by all models, while Schizoid and Avoidant are poorly represented. Overall, GPT-4 emerges as the most effective model in generating results that closely match the human clusters, highlighting its superior performance in profile representation.

Visual representations of these findings can be seen in Figure 2, which shows the clustering of test results according to the personality profiles.  Since davinci is at the time of the experiment considered a legacy model, we can see this by the clustering of evaluated test results and test responses outside the clusters of the baseline. As the GPT model improves with the next iteration of GPT-3.5 and GPT-4, the clusters of evaluated test results and test responses move closer to their respective baseline clusters to the point GPT-4 clusters right above the personality profile location. The degree of closeness within each cluster and the clear separation between clusters further underscores the marked improvement in the performance of the GPT-4 model over the other models.

The violin plots shown in Figure \ref{fig:violin} is a subset of the full report, and provide a clear, visual representation of the distribution of test results for a personality factor. The full violin report can be found in our repository. These plots help understand the variations in results for each LLM and offer a basis for comparison with the human data. Each personality profile's position, indicated by the red circle, intersects with the human's distributions, represented by the green violin. The distributions for davinci and GPT-3.5, represented by the blue and lime violins, respectively, seldom intersect with the actual personality profile, suggesting that these models might not accurately generate test results consistent with the given personality. In contrast, the GPT-4 results intersect with the personality profile within the lower or upper quartile, implying that the model is more successful at generating test results that align closely with the given personality profile. This intersection suggests greater accuracy and consistency with GPT-4, making it a potentially more effective tool for game developers in creating realistic, human-like characters based on the Big Five. These findings underscore the potential of utilizing frontier models, such as GPT-4, to generate more accurate and nuanced character responses in video games, thereby improving the overall gaming experience for players.

\section{Use Case}

Our research shows that frontier LLMs, like GPT-4, can emulate human-like expressions of personality. While this approach could be used within many types of video games, we demonstrate its application using a narrative test scenario from the prototype game, \textit{Dark Shadows} \cite{buongiorno_pangea_2024}. \textit{Dark Shadows} is a detective roleplaying game (RPG) that uses an LLM to generate a mystery for a player to solve. This includes creating the scenario, the solution, important locations, the characters involved, and their dialogue. NPCs are each assigned one of the 20 personality profiles via the 5-tuple vector, representing the Big Five personality. An example case briefing, along with a character description, can be found in Figure \ref{fig:dark_shadows_intro}. 

The influence of personality traits on decision-making processes within the game is similar to how attributes function in role-playing games such as Dungeons and Dragons (DnD). In DnD, characters are defined by attributes like Strength, Dexterity, and Charisma, which influence their ability to perform certain actions or succeed in various challenges. While game designers define the mechanics of these personality traits affecting NPC actions — setting thresholds and determining outcomes based on trait levels — the LLM’s role is to interpret and integrate these traits into the narrative discourse. This collaboration between game mechanics and LLM-driven dialogue generation creates engaging and contextually relevant interactions, enhancing the overall gameplay experience.

To exemplify this feature, we demonstrate the different responses of the same NPCs when changing their personality as the player interrogates them, as shown by Figure \ref{fig:dark_shadows_compare}. Providing the same user input, "\textit{Where were you on the night of the disappearance?},” each response follows their assigned personalities. 

For researchers and practitioners interested in delving deeper into the \textit{Dark Shadows} project, further information can be found through the following resources:

\begin{figure}
  \centering
  \includegraphics[width=\linewidth]{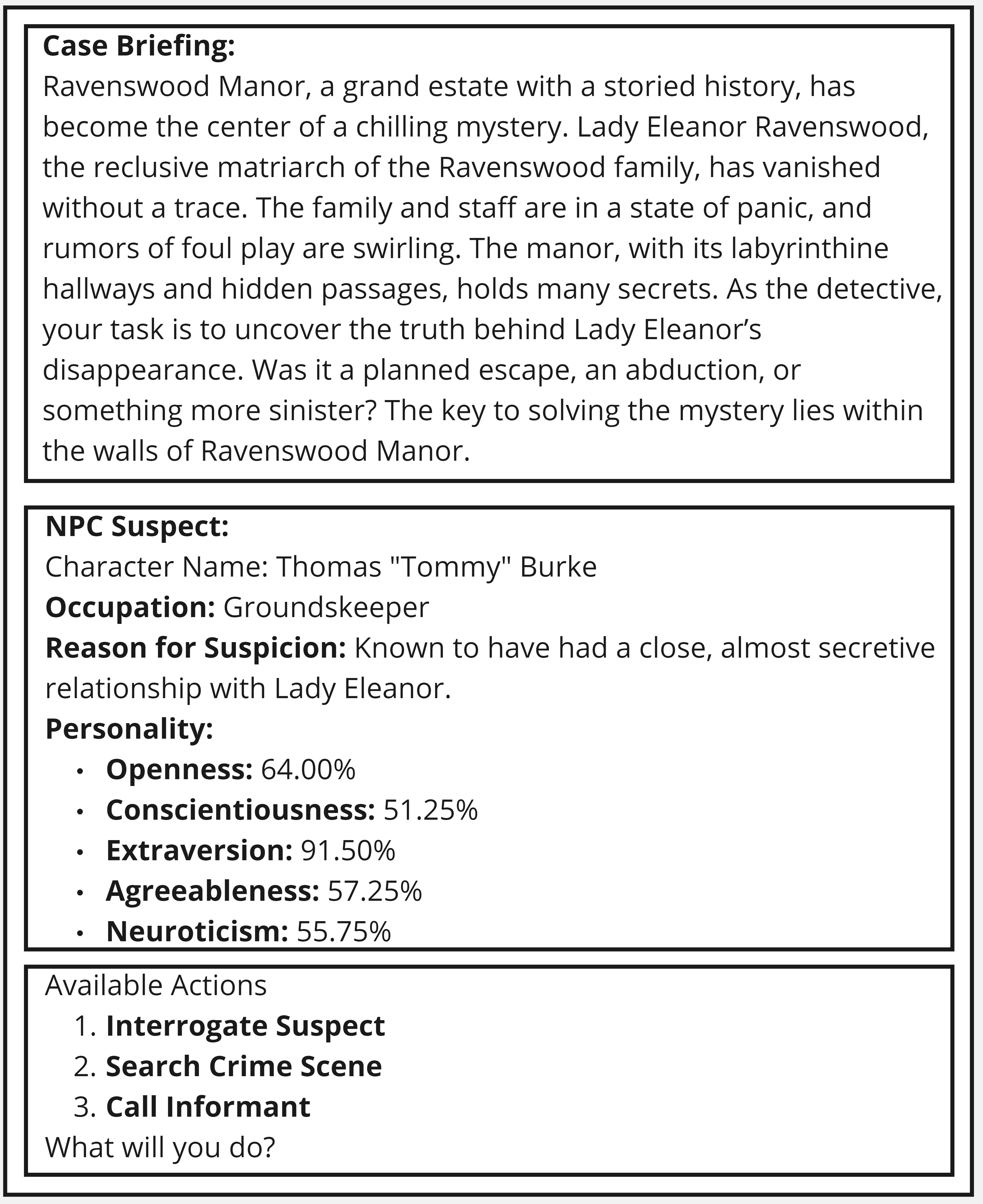}
  \caption{Example case scenario from the prototype game \textit{Dark Shadows}. Only relevant suspects from the case are presented here. A case briefing, followed by the suspect list, and the available actions the player can select.}
  \label{fig:dark_shadows_intro}
\end{figure}

\begin{itemize}
    \item \textbf{Game Trailer}: To see the game in action and understand the practical application of these concepts, watch the \textit{Dark Shadows} game trailer on YouTube {https://youtu.be/H-3-9zxhUlg}.
    \item \textbf{GPT Model Prototype}: Experience a prototype of the \textit{Dark Shadows} game powered by a GPT model. You can try it out {https://chatgpt.com/g/g-RhmfY1KJR-dark-shadows-gpt}.
\end{itemize}

\begin{figure*}
  \centering
  \includegraphics[width=.975\linewidth]{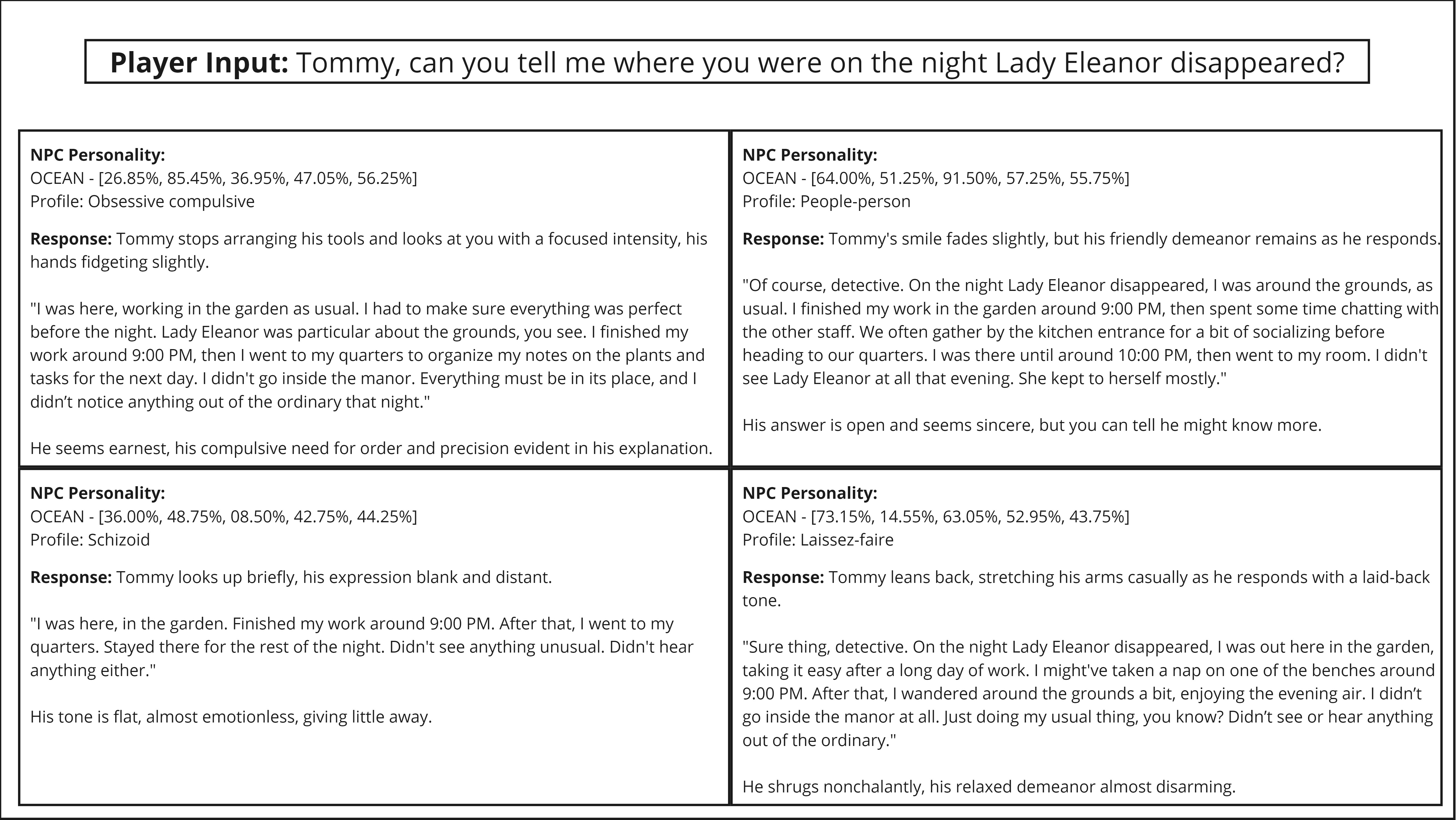}
  \caption{Different responses of a suspect from the game, \textit{Dark Shadows}, when altering their personality but asked the same question.}
  \label{fig:dark_shadows_compare}
\end{figure*}

\section{Limitations}

This section outlines the limitations of our study and identifies areas for future research. While the IPIP-50 was used as the primary tool to assess the Big Five personality traits, it is not the only available method. Future studies should explore how LLMs perform with alternative assessments, such as the IPIP-120 or IPIP-300, to determine whether these tools yield different results in personality emulation. Additionally, examining other personality models like Myers-Briggs or HEXACO, as well as exploring related areas such as emotions, social relationships, or other psychometric variables, could help validate the ability of LLMs to emulate a broader range of human traits.

Our study was limited by the computational and financial resources available. Conducting more extensive tests, such as those using the IPIP-120 or IPIP-300, would require significant resources, especially with large-scale models like GPT-4. While these larger tests may provide deeper insights into LLMs' ability to replicate human personality traits, they were beyond the scope of this research.

Our methodology relied on self-evaluated tests to assess the personality traits generated by LLMs. While this approach is scalable, it limits the model's context awareness. Implementing an interview-style test could allow the model to demonstrate a wider range of behaviors and vocabulary, potentially yielding richer results. However, this would require collaboration with human psychologists and additional resources for response evaluation.

Finally, our study assumes that language models can generate personality-consistent text based on static prompts. However, real-world interactions are dynamic, and ensuring that a model can maintain consistent personality traits over extended, varied interactions remains an open question. Future research should investigate how LLMs can sustain personality coherence across complex and evolving narrative contexts.

These limitations highlight the challenges and opportunities in using LLMs to emulate human personalities in NPCs. Addressing these limitations in future research will help refine the methodologies and improve the accuracy and reliability of personality emulation in digital characters.

\section{Conclusion}
Our investigation into the efficacy of LLMs in emulating realistic human personalities has shown promising potential for creating more human-like NPCs. Using the IPIP-50 questionnaire, we demonstrated that frontier models like GPT-4 can align behaviors with psychometric values, particularly personality traits, resulting in NPCs with consistent and believable behavior patterns. Remarkably, GPT-4 achieved up to 100\% accuracy in emulating certain personality profiles, highlighting its capability to accurately generate human-like responses. By utilizing an AC system that incorporates a personality model, we established a foundational step in demonstrating that LLMs can effectively integrate psychometric variables.

Future research should explore additional psychometric variables, such as NPCs' emotional states, attitudes, and relationships, to further refine their interactions and realism. Fine-tuning LLMs with comprehensive human psychometric datasets will enhance content generation, ensuring that these models can produce dialogue and make decisions that reflect nuanced, human-like behavior. This work emphasizes the need for continued testing to ensure that combining AC systems with LLMs, grounded in human data as a baseline, can create believable, human-like NPCs, thereby unlocking new potentials for depth and realism in in-game characters.

\bibliography{references}

\end{document}